\pdfoutput=1

\documentclass[10pt,twocolumn,letterpaper]{article}

\usepackage{cvpr}
\usepackage{times}
\usepackage{epsfig}
\usepackage{graphicx}
\usepackage{amsmath}
\usepackage{amssymb}

\usepackage[table]{xcolor}
\usepackage{subcaption}
\usepackage{soul}
\usepackage{booktabs}
\usepackage[normalem]{ulem}

\newcolumntype{L}[1]{>{\raggedright\let\newline\\\arraybackslash\hspace{0pt}}m{#1}}
\newcolumntype{C}[1]{>{\centering\let\newline\\\arraybackslash\hspace{0pt}}m{#1}}
\newcolumntype{R}[1]{>{\raggedleft\let\newline\\\arraybackslash\hspace{0pt}}m{#1}}

\newif\ifdraft
\draftfalse
\drafttrue

\definecolor{orange}{rgb}{1,0.5,0}
\definecolor{violet}{RGB}{70,0,170}

\ifdraft
 \newcommand{\PF}[1]{{\color{red}{\bf PF: #1}}}
 \newcommand{\pf}[1]{{\color{red} #1}}
 \newcommand{\MS}[1]{{\color{green}{\bf MS: #1}}}
 \newcommand{\ms}[1]{{\color{green} #1}}
 \newcommand{\ZD}[1]{{\color{violet}{\bf ZD: #1}}}
 
 \newcommand{\YH}[1]{{\color{blue}{\bf YH: #1}}}
 \newcommand{\yh}[1]{{\color{blue}{#1}}}
\else
 \newcommand{\PF}[1]{}
 \newcommand{\pf}[1]{ #1 }
 \newcommand{\KY}[1]{}
 
 \newcommand{\MS}[1]{}
 \newcommand{\ms}[1]{ #1 }
 \newcommand{\ZD}[1]{}
 
 \newcommand{\YH}[1]{}
 \newcommand{\yh}[1]{{#1}}
\fi

\newcommand{\comment}[1]{}
\newcommand{\parag}[1]{\vspace{-3mm}\paragraph{#1}}

\newcommand{\bK}{\mathbf{K}}
\newcommand{\bR}{\mathbf{R}}

\newcommand{\bu}{\mathbf{u}}

\usepackage[pagebackref=true,breaklinks=true,letterpaper=true,colorlinks,bookmarks=false]{hyperref}

\cvprfinalcopy % *** Uncomment this line for the final submission

\def\cvprPaperID{725} % *** Enter the CVPR Paper ID here

\ifcvprfinal\fi
\begin{document}

%%%%%%%%% TITLE
\title{Single-Stage 6D Object Pose Estimation}
%\title{Perspective-n-Cluster for 6D Object Pose Estimation}
%\title{RANSAC-Free 6D Object Pose Estimation}
% \title{Deep Perspective-n-Point for 6D Object Pose Estimation}
% \title{RANSAC-Free 6D Object Pose Estimation}

\author{
	\vspace{0.5em}
	{Yinlin Hu, \quad Pascal Fua, \quad Wei Wang, \quad Mathieu Salzmann} \\
	{CVLab, EPFL, Switzerland} \\
	{\small \{firstname.lastname\}@epfl.ch}\\
}

% \author{%
% 	\vspace{0.5em}
% 	{Yinlin Hu$^1$, \quad Pascal Fua$^1$, \quad Wei Wang$^1$, \quad Vincent Lepetit$^2$, \quad Mathieu Salzmann$^1$} \\
% 	{\small $^1$CVLab, EPFL, Switzerland, \quad $^2$LaBRI, University of Bordeaux, France} \\
% 	{\small \{firstname.lastname\}@epfl.ch, \quad vincent.lepetit@u-bordeaux.fr
% 	, \quad mathieu.salzmann@epfl.ch}\\
% }

\maketitle
\thispagestyle{empty}

% !TEX root = ../top.tex
% !TEX spellcheck = en-US

%%%%%%%%% ABSTRACT
\begin{abstract}

Most recent 6D pose estimation frameworks first rely on a deep network to establish correspondences between 3D object keypoints and 2D image locations and then use a variant of a RANSAC-based Perspective-n-Point (PnP) algorithm. This two-stage process, however, is suboptimal: First,  it is not end-to-end trainable. Second, training the deep network relies on a surrogate loss that does not directly reflect the final 6D pose estimation task.

In this work, we introduce a deep architecture that directly regresses 6D poses from correspondences. It takes as input a group of candidate correspondences for each 3D keypoint and accounts for the fact that the order of the correspondences within each group is irrelevant, while the order of the groups, that is, of the 3D keypoints, is fixed. Our architecture is generic and can thus be exploited in conjunction with existing correspondence-extraction networks so as to yield single-stage 6D pose estimation frameworks. Our experiments demonstrate that these single-stage frameworks consistently outperform their two-stage counterparts in terms of both accuracy and speed.

\end{abstract}
% !TEX root = ../top.tex
% !TEX spellcheck = en-US

\section{Introduction}
\label{sec:introduction}

Detecting 3D objects in images and computing their 6D pose must be addressed in a wide range of applications~\cite{Hinterstoisser12b,Michel17,Wang19d,Maji09a}, ranging from robotics to augmented reality. State-of-the-art approaches~\cite{Rad17,Tekin18a,Oberweger18,Jafari18,Hu19a,Peng19a,Zakharov19a,Park19,Li19a} follow a two-stage paradigm: First use a deep network to establish correspondences between 3D object points and their 2D image projections, then use a RANSAC-based  Perspective-n-Point (PnP) algorithm to compute the 6 pose parameters~\cite{Hartley00,Lepetit05b,Rothganger06,Wagner08,Lepetit09,Kneip14,Ferraz14,Urban16}. 

\begin{figure}[t]
    \begin{center}
    \begin{tabular}{cc}
    \includegraphics[width=0.45\linewidth, clip, trim=100 10 80 20]{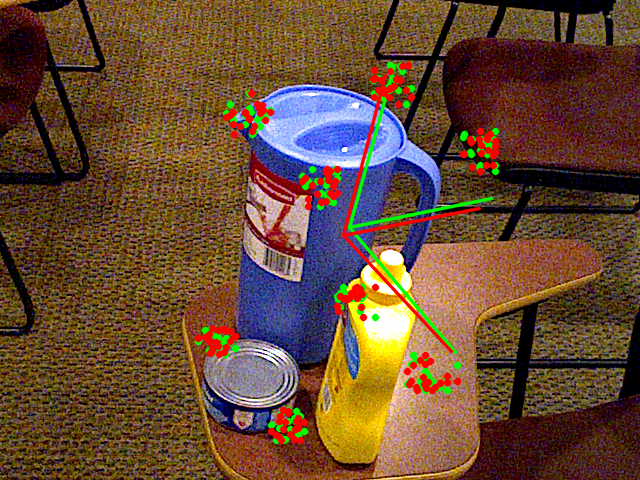} &
    \includegraphics[width=0.45\linewidth, clip, trim=100 10 80 20]{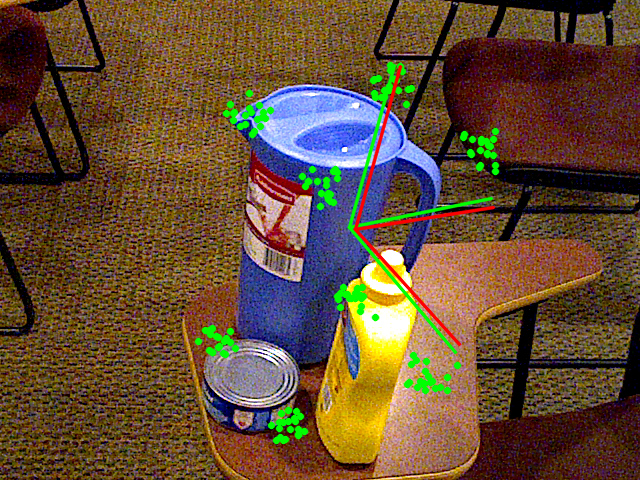} \\
    (a) & (b) \\
    \end{tabular}
    % \fbox{\rule{0pt}{1in} \rule{0.25\linewidth}{0pt}}
    \end{center}
    \vspace{-6mm}
    \caption{{\bf Motivation.} Consider the modern 6D pose estimation algorithm of~\cite{Hu19a} that uses a deep network to predict several 2D correspondences for each of the eight 3D corners of the pitcher's bounding box. {\bf (a)} Because it minimizes the average 2D error of these correspondences, two instances of such a framework could produce correspondences that differ but have the same {\em average} accuracy, such as the green and the red ones. As evidenced by the projected green and red reference frames, applying a RANSAC-based PnP algorithm to these two sets of correspondences can yield substantially different poses. {\bf (b)} Even when using only the set of green correspondences, simply changing their order causes a RANSAC-based PnP algorithm to return different solutions.    }
    \label{fig:motivation}
\end{figure}

While effective, this paradigm suffers from several weaknesses. First, the loss function used to train the deep network does not reflect the true goal of pose estimation, but encodes a surrogate task, such as minimizing the 2D errors of the detected image projections. The relationship between such errors and the pose accuracy, however, is not one-to-one.  As shown in Fig.~\ref{fig:motivation}~(a) for the state-of-the-art framework of~\cite{Hu19a}, two sets of correspondences with the same {\it average} 2D error can result in different pose estimates. Second, the two-stage process is not end-to-end trainable. Finally, the iterative RANSAC is time-consuming when there are many correspondences that need to be handled.
% Second, the correspondences are established individually. This fails to exploit the fact that knowing the location of the 2D projection of one of the 3D points imposes constraints on the potential locations of the others' projections.  Finally the two-stage process is not end-to-end trainable.

% !TEX root = ../top.tex
% !TEX spellcheck = en-US

In principle, an end-to-end framework could be designed by exploiting a deep version of RANSAC~\cite{Brachmann16b,Brachmann19}, followed by another network performing pose estimation from correspondences~\cite{Dang18a}. However, the time-consuming character of RANSAC in the presence of many outliers, and the poor repeatability of its solution, arising from the fact that, as shown in Fig.~\ref{fig:motivation}~(b), the order of the correspondences affects the resulting pose, do not make it a good candidate for inclusion into an end-to-end trainable network. Furthermore, the approach of~\cite{Dang18a} relies on using a Direct Linear Transform (DLT)~\cite{Hartley00} to compute the pose, which is known to be imprecise and would exacerbate in the poor repeatability problem.

As a result, there are still no end-to-end frameworks that can handle jointly keypoint localization and 6D pose estimation. In this paper, we overcome this by introducing a simple but effective network that directly regresses the 6D pose from groups of 3D-to-2D correspondences associated to each 3D object keypoint. Its architecture explicitly encodes that the order of the correspondences in each group is irrelevant, while exploiting the fact that the order of the groups is fixed and corresponds to that of the 3D keypoints.

We then demonstrate the generality of this network by combining it with two state-of-the-art  correspondence-extraction frameworks~\cite{Hu19a,Peng19a}. This yields end-to-end trainable 6D pose estimation frameworks that are both accurate and repeatable.
We show that these single-stage frameworks systematically outperform the original two-stage ones~\cite{Hu19a,Peng19a}, in terms of both accuracy and runtime.

\section{Related Work}
\label{sec:related}

Detecting keypoints in the input image followed by running a RANSAC-based PnP algorithm on the established 3D-to-2D correspondences is a classical way for solving the 6D object pose estimation problem.
Over the years, many methods have been proposed to improve 3D-to-2D matching~\cite{Lowe04,Tola10,Trzcinski12c,Tulsiani15,Pavlakos17a,Ono18}, relying on diverse techniques, such as template-matching~\cite{Hinterstoisser12a,Hinterstoisser12b}, edge-matching~\cite{li11,Lowe91}, and 3D model-based matching~\cite{Huttenlocher93,liu10a,Hsiao14a}. However, these traditional methods still often fail in the presence of severe occlusions and cluttered background.

As in many other areas, the modern take on 6D object pose estimation from an RGB image involves deep neural networks. The simplest approach is to directly regress from the image to the pose parameters~\cite{Kehl17,Xiang18b}. However, this tends to be less accurate than first  establishing 3D-to-2D correspondences~\cite{Rad17,Tekin18a,Oberweger18,Jafari18,Hu19a,Peng19a,Zakharov19a,Park19,Li19a} and then running a RANSAC-based Perspective-n-Point (PnP) algorithm~\cite{Hartley00} to estimate the object position and orientation given the camera intrinsic parameters. What these methods all have in common is that the correspondences are established independently from each other and consistency is only imposed after the fact by the RANSAC PnP algorithm, which is not part of the deep network. As shown in~\cite{Yi16a}, albeit in a different context, this fails to exploit the fact that all correspondences are constrained by the camera pose and are therefore {\it not} independent from each other.

Our goal in this paper is to turn the two-stage process described above into a single-stage one by implementing the RANSAC-based PnP part of the process as a deep network that can be combined with the one that establishes the correspondences. This is not a trivial problem because the standard approach to PnP involves performing a Singular Value Decomposition (SVD), which can be embedded in a deep network but often results in numerical instabilities. In~\cite{Dang18a}, this was addressed by avoiding the explicit use of SVD and instead treating PnP as a least-square fitting problem via the Direct Linear Transform (DLT) approach~\cite{Hartley00}. This, however, does not guarantee that the result describes a true rotation and further post processing is still needed. 

By contrast, the backpropagation-friendly eigendecomposition method of~\cite{Wang19b} performs explicit SVD, and could in principle used to perform PnP. Doing so, however, would fail to account for the RANSAC part of the algorithm to select the correct correspondences.
While RANSAC can be implemented via a deep network~\cite{Brachmann16b,Brachmann19}, its poor repeatability, evidenced in Fig.~\ref{fig:motivation}(b), makes it ill-suited to train an end-to-end 6D pose estimation network.  In short, no one yet has proposed a satisfying solution to designing a single-stage 6D pose estimation network, which is the problem we address here.

Our architecture is inspired by PointNet~\cite{Qi17a,Qi17b}. However, PointNet was designed to deliver invariance to rigid transformations, which is the opposite of what we need. Furthermore, we introduce a grouped feature aggregation scheme to effectively hande correspondence clusters in 6D object pose estimation.

%-------
% OLD
%--------

\comment{
To do this, we develop a PnP-like network inspired by the PointNet~\cite{Qi17a,Qi17b} architecture and its simplified but effective variant used in~\cite{Yi18a}. However, instead of using the architecture of~\cite{Yi18a}, which we found to be too cumbersome for our purposes, we design a more effective one that inherently accounts for the properties of our 3D-to-2D correspondences, in particular the fact that they form groups corresponding to the individual keypoints, and that, while the order within each group is irrelevant, the order of the groups is fixed because the set of 3D keypoints is given.

Note that, while here we focus on 6D pose estimation from RGB images, we believe that our PnP-like network could easily be incorporated into RGBD-based 6D pose estimation frameworks~\cite{Hinterstoisser12b,Brachmann14,Brachmann16a,Michel17,Wang19d}, which also typically follow a two-stage process.} 

% !TEX root = ../top.tex
% !TEX spellcheck = en-US

\section{Approach}
\label{sec:approach}

Given an RGB image captured by a calibrated camera,  our goal is to simultaneously detect objects and estimate their 6D pose. We assume them to be rigid and their 3D model to be available.  In this section, we first formalize the 6D pose estimation problem assuming that sets of 2D correspondence are given {\it a priori} for each 3D keypoint on the target object and propose a network architecture that yields 6D poses from such inputs. This network is depicted by Fig.~\ref{fig:arch}.  We then discuss how to obtain a single-stage 6D pose estimation framework when these correspondences are the output of another network.

\subsection{6D Pose from Correspondence Clusters}
\label{sec:problem_formulation}

\begin{figure}[t]
    \begin{center}
    \begin{tabular}{cc}
    \includegraphics[width=0.48\linewidth]{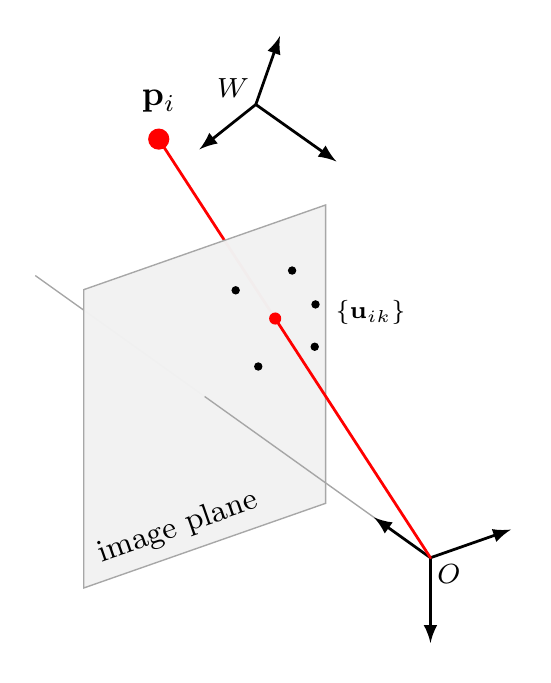} &
    \includegraphics[width=0.48\linewidth]{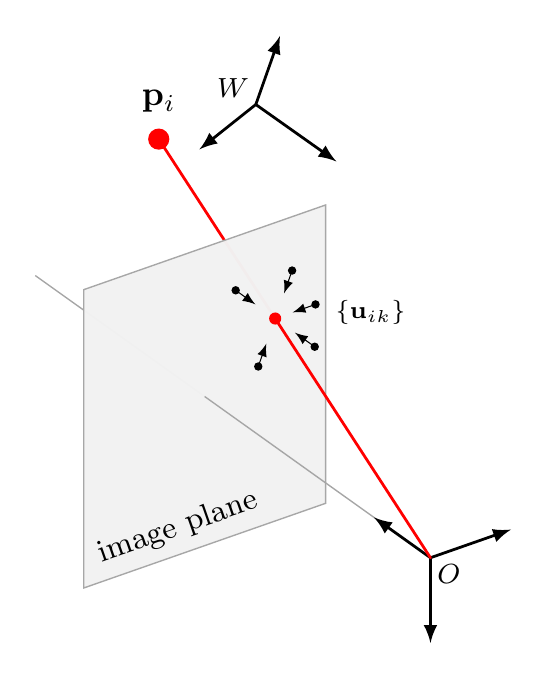} \\
    (a) &  (b) \\
    \end{tabular}
    % \fbox{\rule{0pt}{1in} \rule{0.25\linewidth}{0pt}}
    % \includegraphics[width=0.8\linewidth]{./fig/geometry/geometry2.pdf}
    \end{center}
    \vspace{-6mm}
    \caption{{\bf 3D to 2D correspondences.}
    %Problem formulation 
    {\bf (a)} Given $m$ potential 2D correspondences $\bu_{ik}$ for each one of $n$ 3D object keypoints ${\bf p}_i$, $\{({\bf p}_i \leftrightarrow \bu_{ik})\}_{1 \leq i \leq n, 1 \leq k \leq m}$, the pose can be computed based on these 3D-to-2D correspondences. Here, we only show the correspondence cluster for ${\bf p}_i$. The camera and object coordinate systems are denoted by $O$ and $W$ respectively. {\bf (b)} The pose can also be obtained from point-to-vector correspondences, in which case a 3D-to-2D correspondence is defined between a 3D point and a 2D vector. Our method can handle both cases.
    }
    \label{fig:geometry_demo}
\end{figure}

Let us assume that we are given the $3\times3$ camera intrinsic parameter matrix ${\bf K}$ and $m$ potential 2D correspondences $\bu_{ik}$ for each one of $n$ 3D object keypoints ${\bf p}_i$, with $1 \leq i \leq n$ and  $1 \leq k \leq m$. The ${\bf p}_i$ is expressed in a coordinate system linked to the object, as shown in Fig.~\ref{fig:geometry_demo}(a). For each {\it valid} 3D to 2D correspondence ${\bf p}_i \leftrightarrow \bu_{ik}$, we have 
\begin{equation}
\begin{aligned}
\lambda_{ik}
\begin{bmatrix}
{\bf u}_{ik} \\
1 \\
\end{bmatrix}
=\bK(\bR{\bf p}_i+{\bf t}),
\end{aligned}
\label{eq:perspective}
\end{equation}
where $\lambda_i$ is a scale factor, and $\bR$ and ${\bf t}$ are the rotation matrix and translation vector that define the camera pose. Because ${\bf R}$ is a rotation, it only has three degrees of freedom and  ${\bf t}$ likewise, for a total of 6.

Note that the 3D-to-2D correspondences above are not restricted to 3D point to 2D point correspondences. In particular, as shown in Fig.~\ref{fig:geometry_demo}(b), our formalism can handle 3D point to 2D \emph{vector} correspondences, which have been shown to be better-suited to use in conjunction with a deep network~\cite{Peng19a}. In that case, the 2D locations can be infered as the crosspoint of two 2D vectors, and Eq.~\ref{eq:perspective} still holds on crosspoints. Our approach as discussed below also still applies, and we therefore do not explicitly distinguish between these two types of 3D-to-2D correspondences unless necessary.

Classical PnP methods~\cite{Lepetit09,Ferraz14,Urban16} try to recover $\bf R$ and ${\bf t}$ given several correspondences, which typically involves using RANSAC to find the valid ones. In the process, an SVD has to be performed on the many randomly chosen subsets of correspondences that must be tried before one containing only valid correspondences is found. In this work, we propose to replace this cumbersome process by a non-linear regression implemented by an appropriately designed deep network $g$ with parameters $\Theta$. In other words, we have
\begin{align}
    ({\bf R},{\bf t}) &= g(\{({\bf p}_i  \leftrightarrow {\bf u}_{ik})\}_{1 \leq i \leq n,1 \leq k \leq m};\Theta) \; .          
    \label{eq:regressor}
\end{align}

We now turn to the actual implementation of $g_{\theta}$. In the remainder of this section, we first discuss the properties of the set of 3D to 2D correspondences ${\cal C}_2^3 =\{({\bf p}_i  \leftrightarrow {\bf u}_{ik})\}_{1 \leq i \leq n,1 \leq k \leq m}$ that the network takes as input and then the architecture we designed to account for them. 

\subsubsection{Properties of the Correspondence Set}

We will refer to all the 2D points associated to a specific 3D point as a {\it cluster} because, assuming that the algorithm used to find them is a good one, they tend to cluster around the true location of the 3D point's projection, as can be seen in Fig.~\ref{fig:motivation}. Our implementation choice were driven by the following considerations:

{\bf Cluster ordering.} 
The order of the correspondences within a cluster is irrelevant and should not affect the result. However, the order of the clusters corresponds to the order of the 3D points, which is given and fixed.

{\bf Interaction within a cluster and across clusters.}
Although the points in the same cluster correspond to the same 3D point, the 2D location estimate for each point should be expected to be noisy. Thus the model needs to capture the noise distribution within each cluster. More importantly, one single cluster can tell us nothing about the pose, and the final pose can only be inferred by capturing the global structure for multiple clusters. 

{\bf Rigid transformations matter.} 
When processing 3D point clouds with a deep network, one usually wants the result to be invariant to rigid transformations. By contrast, here, we want our 2D points to represent projections of 3D points, and the features that we extract from them should depend on their absolute positions, which are critical to pose estimation.

% !TEX root = ../top.tex
% !TEX spellcheck = en-US

% !TEX root = ../top.tex
% !TEX spellcheck = en-US

\begin{figure*}[t]
    \begin{center}
    \includegraphics[width=0.99\linewidth]{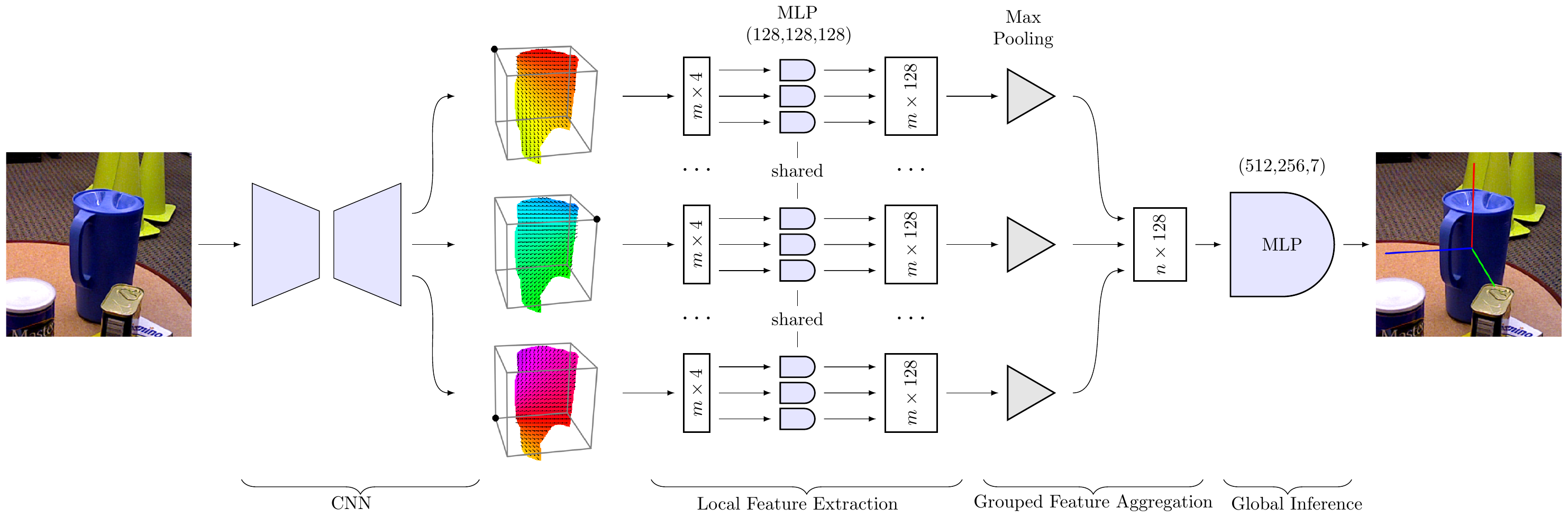}
    \end{center}
    \vspace{-6mm}
    \caption{{\bf Overall architecture for single-stage 6D object pose estimation.}
    After establishing 3D-to-2D correspondences by some segmentation-driven CNN for 6D pose~\cite{Hu19a,Peng19a}, we use three main modules to infer the pose from these correspondence clusters directly: a local feature extraction module with shared network parameters, a feature aggregation module operating
    % in groups according to different clusters, 
    within the different clusters, and a global inference module consisting of simple fully-connected layers to estimate the final pose as a quaternion and a translation. The color in the CNN outputs indicates the direction of the 2D offset from the grid cell center to the corresponding projected 3D bounding box corner.
    }
    \label{fig:arch}
\end{figure*}

\subsubsection{Network Architecture}
\label{sec:network}

We construct a simple network architecture, depicted by Fig.~\ref{fig:arch}, that utilizes the properties discussed above to predict the pose from correspondence clusters. It comprises three main modules: A local feature extraction module with shared network parameters, a feature aggregation module within individual clusters, and a global inference module made of simple fully-connected layers.

{\bf Local feature extraction.}
We use an MLP with three
% \ms{fully-connected} 
layers to extract local features for each correspondence, with weights shared across the correspondences and across the clusters.

{\bf Grouped feature aggregation.}
As the order of the clusters is given but the points within each cluster are orderless, to extract the representation for each cluster, we design a grouped feature aggregation method that insensitive to the correspondence order. In theory, we could have used an architecture similar to that of PointNet~\cite{Qi17a,Qi17b}. However, PointNet is designed to deliver invariance to rigid transformations, which is the opposite of what we need. Instead, given $n$ clusters, each containing $m$ 2D points $\{{\bf u}_{ik}\}, 1 \leq i \leq n, 1 \leq k \leq m$, we define a set function $\mathcal{F}:\mathcal{X}\rightarrow\mathbb{R}^{nD}$ that maps correspondences $\{{\bf u}_{ik}\}_{1 \leq k \leq m}$ to the $nD$-dimensional vector 
\begin{small}
\begin{equation}
\begin{split}
    \displaystyle
    CAT\Big(\underset{k}{MAX}(\{{\bf f}_{1k}\}), \underset{k}{MAX}(\{{\bf f}_{2k}\}), .. , \underset{k}{MAX}(\{{\bf f}_{nk}\})\Big),
\end{split}
\label{eq:pooling}
\end{equation}
\end{small}
where ${\bf f}_{ik}$ is the $D$-dimensional feature representation of ${\bf u}_{ik}$ obtained via the above-mentioned fully-connected layers, $MAX()$ is the max-pooling operation and $CAT()$ is the concatenation operation. In our experiments, we found that neither instance normalization~\cite{Ulyanov16a,Yi18a} nor batch normalization~\cite{Ioffe15} improved the performance here. Therefore, we do not use these operations in our network $g_{\theta}$. 
%\PF{Should we say something about~\cite{Yi18a} at this point. They normalize and you don't?} 

In principle, one could use a single max-pooling operation, without accounting for the order of the groups, just as PointNet~\cite{Qi17a} does to achieve permutation invariance for all points. In our case, however, this would mean ignoring the property that the order of the groups is fixed. By contrast, Eq.~\ref{eq:pooling} is invariant to any permutation within a cluster but still accounts for the pre-defined cluster order. We demonstrate the benefits of this approach in the results section. 

{\bf Global inference.}
We then pass the $nD$-dimensional vector aggregating the group features through another MLP which outputs the 6D pose. To this end, we use three fully-connected layers and encode the final pose as a quaternion and a translation.

\subsection{Single-Stage 6D Object Pose Estimation}
\label{sec:single-stage}

The deep network described above gives us a differentiable way to predict the 6D pose from correspondence clusters for a given object.
Given the input image, we therefore still need to detect each object and establish the 3D to 2D correspondences. To do so, we use another deep regressor $f$ with parameters $\Phi$, which, for one object, lets us write
\begin{equation}
    \left[{\bf u}_{i1},\ldots,{\bf u}_{im}\right]=f({\bf p}_i, {\bf I};\Phi), \quad 1 \leq i \leq n
\label{eq:local_prediction}
\end{equation}
where ${\bf I}$ is the input RGB image. To implement $f$, we use the recent encoder-decoder architecture of either~\cite{Hu19a} or~\cite{Peng19a}.

In practice, the $\{{\bf p}_i\}$ are often taken to be the eight corners of the 3D bounding box of the object's 3D model~\cite{Rad17,Oberweger18,Hu19a}, which leads to different 3D points $\{{\bf p}_i\}$, for different object types. In our experiments, we  have observed that using the same $\{{\bf p}_i\}$ for every object has little impact on the accuracy of $f_{\phi}$ and makes the subsequent training of $g_{\theta}$ much easier. We therefore use a single cube for all dataset objects, defined as the largest cube contained by a sphere whose radius is the average of that of the bounding spheres of all object 3D models. This means that the 3D keypoint coordinates are implicitly given by the order of the clusters and do not need to be explicitly specified as network inputs. We therefore a use of 4D representation for each input correspondence, which does {\it not} include the 3D coordinates. Instead, because the network of~\cite{Hu19a} operates on an image grid, when we use it to find the correspondences, we take the input to be the $x$ and $y$ coordinates of the center of the grid cell in which the 2D projections are and the $dx$ and $dy$ offsets from that center. In other words, the image coordinates of a 2D correspondence are $x+dx$ and $y+dy$. We tried using these directly as input but we found out experimentally that giving the network what amounts to a first order expansion works better.  When using the network of~\cite{Peng19a} instead of that of~\cite{Hu19a} to find the correspondences, we use the same input format but normalize the $dx$ and $dy$ so that they represent an orientation.

Our complete model can therefore be written as
\begin{align}
    ({\bf R},{\bf t}) &= g\big(f({\bf p}_1, {\bf I};\Phi),\cdots,f({\bf p}_8, {\bf I};\Phi); \Theta\big) \; .          \label{eq:singlestate}
\end{align}
To train it, we minimize the loss function
\begin{equation}
{\mathcal L}={\mathcal L}_{s} + {\mathcal L}_{k} + {\mathcal L}_{p}\;,
\label{eq:lossF}
\end{equation}
which combines segmentation term ${\mathcal L}_{s}$ aiming to assign each grid cell to an object class of to the background, a keypoint regression term ${\mathcal L}_{k}$, and a pose estimation term ${\mathcal L}_{p}$. 
%\PF{I don't understand what "segmentation term" means.} \YH{It is used to extract visible parts of each instance, and our method works only on correspondences extracted from grids within visible parts. Maybe should we change it to ``category-labeling term''?} 
We take ${\mathcal L}_{s}$ to be the Focal Loss of~\cite{Lin17}, and  ${\mathcal L}_{k}$ to be the regression term of either~\cite{Hu19a} or~\cite{Peng19a} depending on which of the two architectures we use. As in~\cite{Xiang18b,Li18a}, we take ${\mathcal L}_{p}$ to be the 3D space reconstruction error, that is
\begin{equation}
    {\mathcal L}_{p} = \frac{1}{n}\sum_{i=1}^{n} \Vert (\hat{\bf R}{\bf p}_i+\hat{\bf t})-({\bf R}{\bf p}_i+{\bf t}) \Vert \;,
    \label{equ:pose_term}
\end{equation}
where $\hat{\bf R}$ and $\hat{\bf t}$ are the estimated rotation matrix and translation vector, ${\bf R}$ and ${\bf t}$ are the ground-truth ones. The rotations are estimated from the estimated and ground-truth quaternions, which can be done in a differentiable manner~\cite{Zheng13}. We also normalize the translations to make sure the regression targets all have a comparable range.

%\PF{In my own experiments, I directly used the $L^2$ distance between quaternions. Why is this one better?} \YH{I think $L^2$ distance between quaternions is ok for rotation supervision, while it is not easy to balance the relative sum weight between rotation and translation if we want to report pose accuracy in one number. And also, nearly all SoA methods use 3D space reconstruction error to report the pose accuracy on benchmarks, so I directly optimize this target.}

Our architecture simultaneously outputs a segmentation mask and potential 2D locations for a set of predefined 3D keypoints.  More specifically, for a dataset with $S$ object classes and an input image ${\bf I}$ of size $h \times w \times 3$, it outputs a 3D tensor of size $H \times W \times C$. The dimensions $H$ and $W$ are proportional to the input resolution and $C=(S+1)+2*n$ with $(S+1)$ channels for segmentation, including one for the background class, and $2*n$ for the 2D locations (or 2D direction vectors) corresponding to the $n$ 3D points ${\bf p}_i$. To obtain correspondence clusters for a given object, we randomly sample $m=200$ grid cells on the output feature tensor that fall under the segmentation mask of a particular class label.

\section{Experiments}
\label{sec:experiments}

% !TEX root = ../top.tex
% !TEX spellcheck = en-US

\begin{figure}[t]
    \begin{center}
    \includegraphics[width=0.24\linewidth,clip, trim=2 2 2 2]{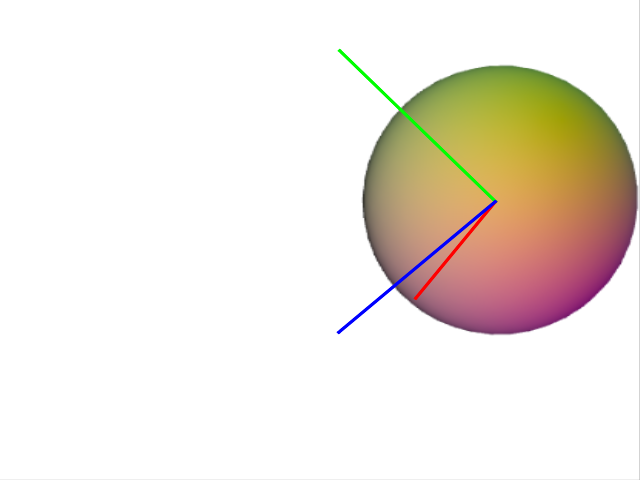}
    \includegraphics[width=0.24\linewidth,clip, trim=2 2 2 2]{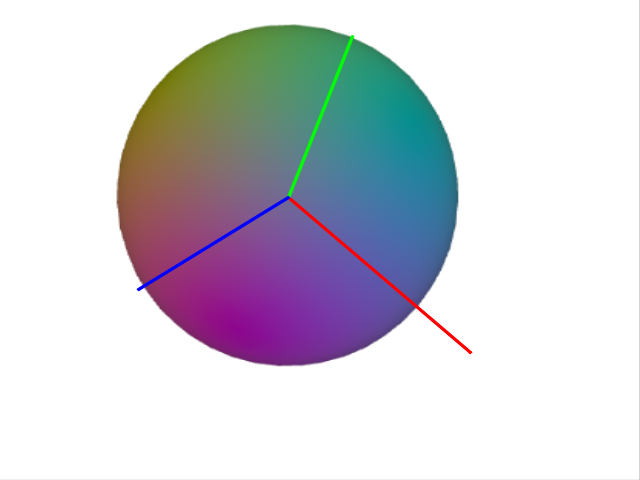}
    \includegraphics[width=0.24\linewidth,clip, trim=2 2 2 2]{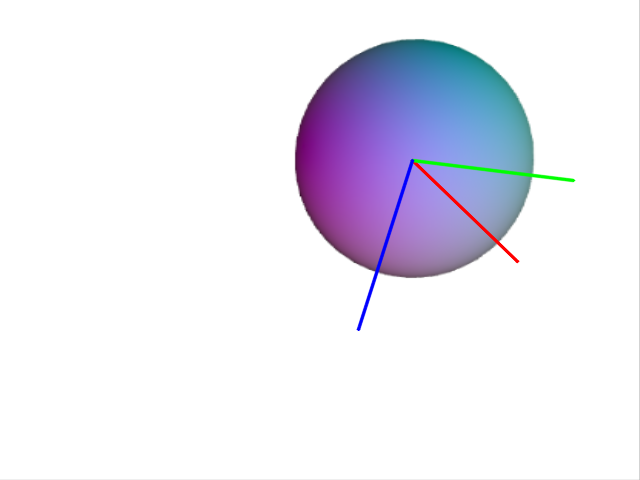}
    \includegraphics[width=0.24\linewidth,clip, trim=2 2 2 2]{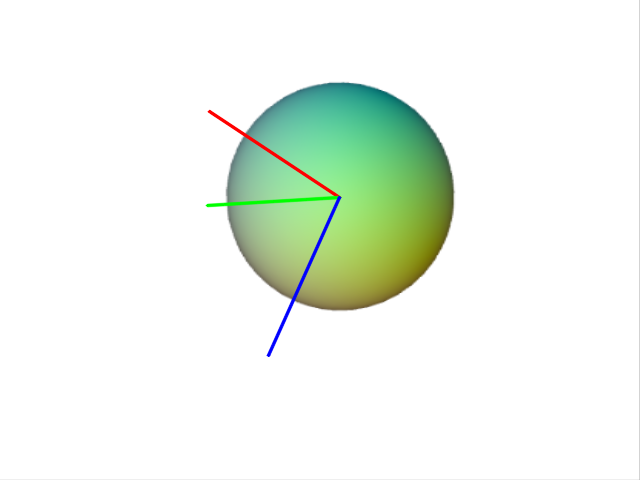}
    \end{center}
    \vspace{-6mm}
    \caption{{\bf Synthetic data.} We create synthetic data by randomly changing the pose of a unit sphere in 3D space relative to the camera. We capture 20K images for training and 2K for testing.}
    \label{fig:synthetic_demo1}
\end{figure}

We compare our single-stage approach to more traditional but state-of-the-art two-stage frameworks~\cite{Hu19a,Peng19a}, first on synthetic data and then on real data from the challenging Occluded-LINEMOD~\cite{Krull15} and YCB-Video~\cite{Xiang18b} datasets. Our source code is publicly available at \href{https://github.com/cvlab-epfl/single-stage-pose}{https://github.com/cvlab-epfl/single-stage-pose}.

\subsection{Synthetic Data}
\label{sec:eval_synthetic}

% !TEX root = ../top.tex
% !TEX spellcheck = en-US

\begin{figure}[t]
    \begin{center}
    \includegraphics[width=0.24\linewidth,clip, trim=100 90 90 0]{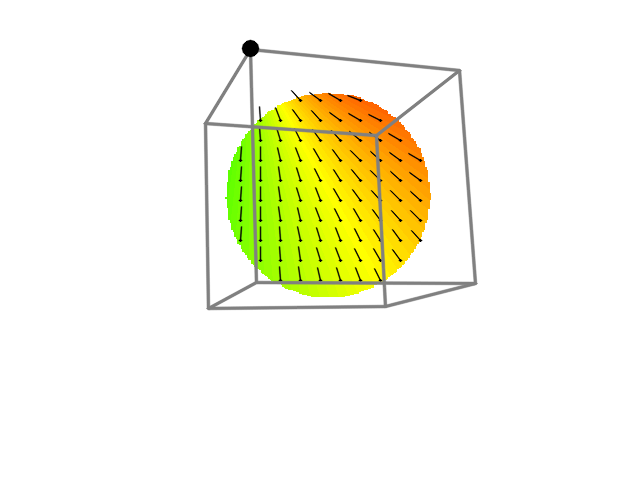}
    \includegraphics[width=0.24\linewidth,clip, trim=100 90 90 0]{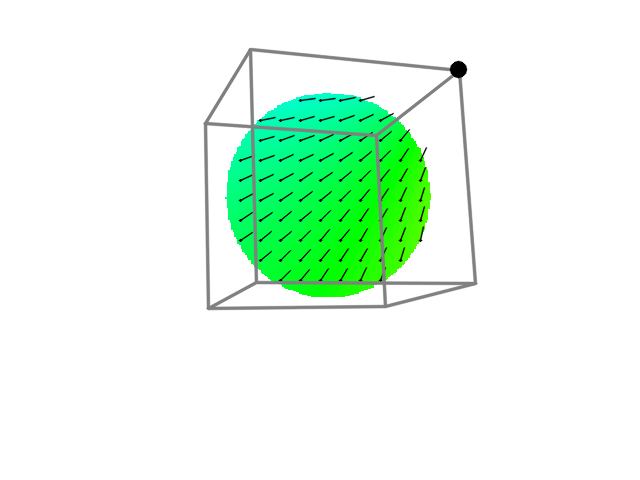}
    \includegraphics[width=0.24\linewidth,clip, trim=100 90 90 0]{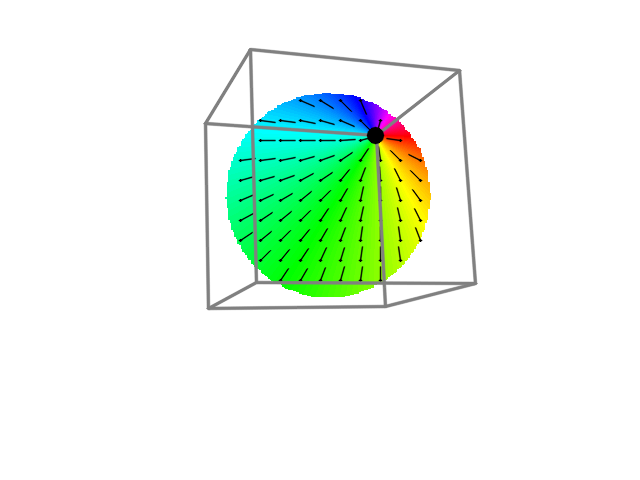}
    \includegraphics[width=0.24\linewidth,clip, trim=100 90 90 0]{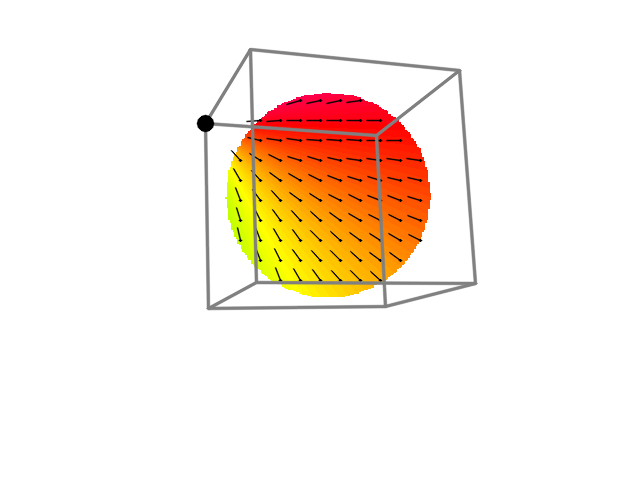}
    \\
    \vspace{1em}
    \includegraphics[width=0.24\linewidth,clip, trim=100 90 90 0]{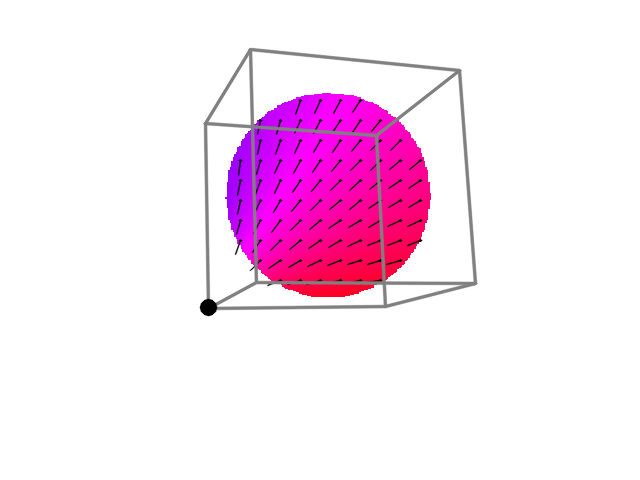}
    \includegraphics[width=0.24\linewidth,clip, trim=100 90 90 0]{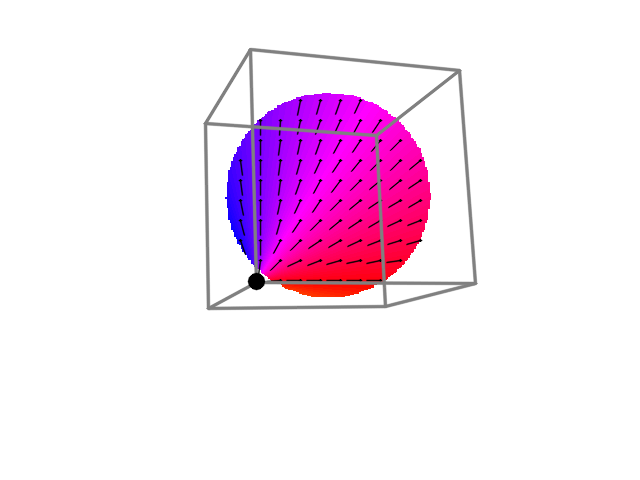}
    \includegraphics[width=0.24\linewidth,clip, trim=100 90 90 0]{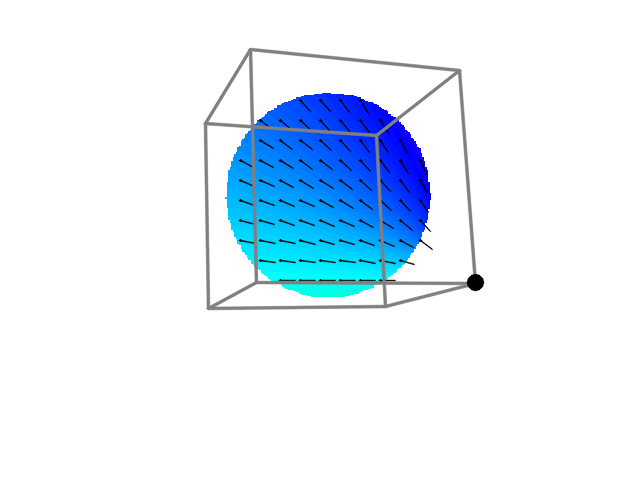}
    \includegraphics[width=0.24\linewidth,clip, trim=100 90 90 0]{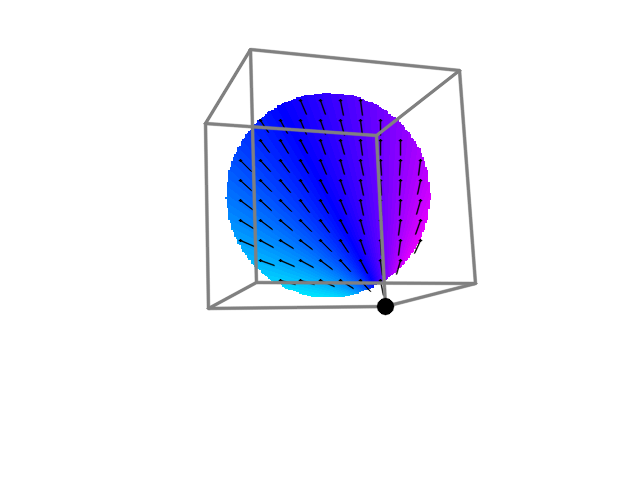}
    \end{center}
    \vspace{-6mm}
    \caption{{\bf Generating correspondences.} We project each corner of the sphere's 3D bounding box in the image and, for each grid cell within the object mask, create a correspondence by recording the center $x,y$ of the grid cell and the offset $dx,dy$ to the projected corner.
    %Given the eight corners of the 3D bounding box of the sphere, we let each pixel within the mask connecting the corresponding 3D point by a 4-dim vector $(x,y,dx,dy)$, where $x$ and $y$ are the 2D position coordinates and $dx$ and $dy$ are the offsets to the corresponding 2D reprojection.
    }
    \label{fig:synthetic_demo2}
\end{figure}

As in~\cite{Lepetit09, Ferraz14}, we create synthetic 3D-to-2D correspondences using a virtual calibrated camera, with image size $640\times 480$, focal length $800$,  and principal point at the image center. We take our target object to be a unit 3D sphere, which we randomly rotate and whose center we randomly translate within the interval $[-2,2]\times[-2,2]\times[4,8]$ expressed in the camera coordinate system, as shown in Fig.~\ref{fig:synthetic_demo1}. 

Recall from Section~\ref{sec:single-stage}, that $g_{\theta}$, the network that regresses poses from the correspondence clusters, expects 4D inputs in the form $[x,y,dx,dy]$, where $x,y$ represent the center of an image grid location and $dx,dy$ a shift from that center. Here, each one should represent a potential image correspondence for a specific corner of the sphere's bounding box for a particular object. Given the segmentation mask of a particular object obtained by projecting the object's 3D model in the image, we create correspondences in the following manner. We project each corner of the sphere's 3D bounding box in the image and, for each grid cell in the segmentation mask, record the cell center $x,y$ and the displacement $dx,dy$ to the projected corner. We then take the resulting correspondences from 200 randomly sampled grid cells within the mask. We add Gaussian noise to their $dx,dy$ values as well as create outliers by setting some percentage of the $dx,dy$ to values uniformly sampled in the image. Fig.~\ref{fig:synthetic_demo2} demonstrates this procedure.

We trained $g_{\theta}$ for 300 epoch on 20K synthetic training images with batch size 32, and a learning rate of 1e-3 using the Adam optimizer. During training we randomly add 2D noise with variance $\sigma$ in the range of [0, 15] and create from 
0\% to 30\% of outliers. To test the accuracy obtained with different noise levels and outlier rates, we use 2K synthetic test images and report the mean pose accuracy in terms of the ratio of the 3D space reconstruction error of Eq.~\ref{equ:pose_term} to the diameter of the target object.

\begin{figure}[t]
    \begin{center}
    \includegraphics[width=0.49\linewidth]{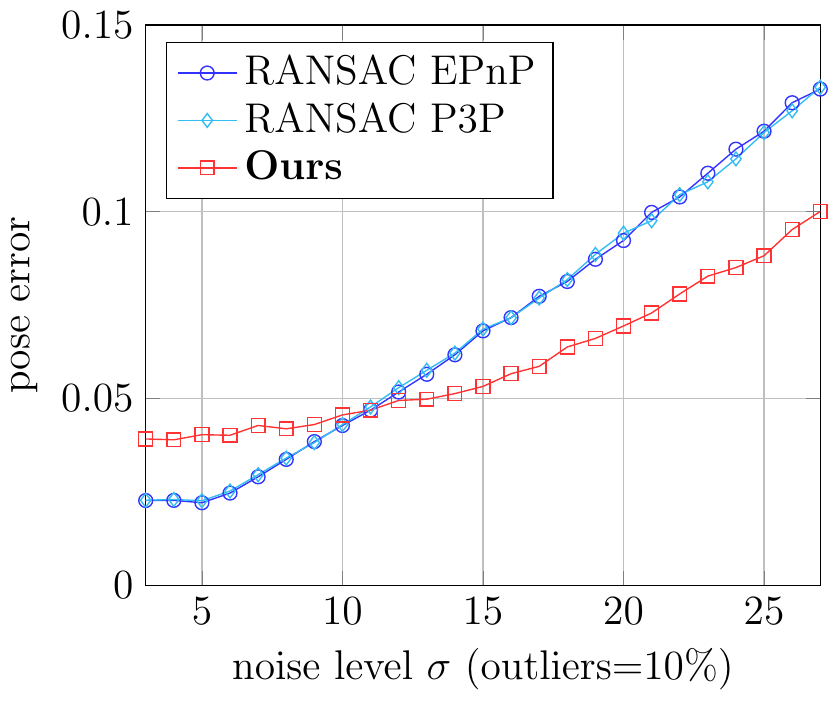}
    \includegraphics[width=0.49\linewidth]{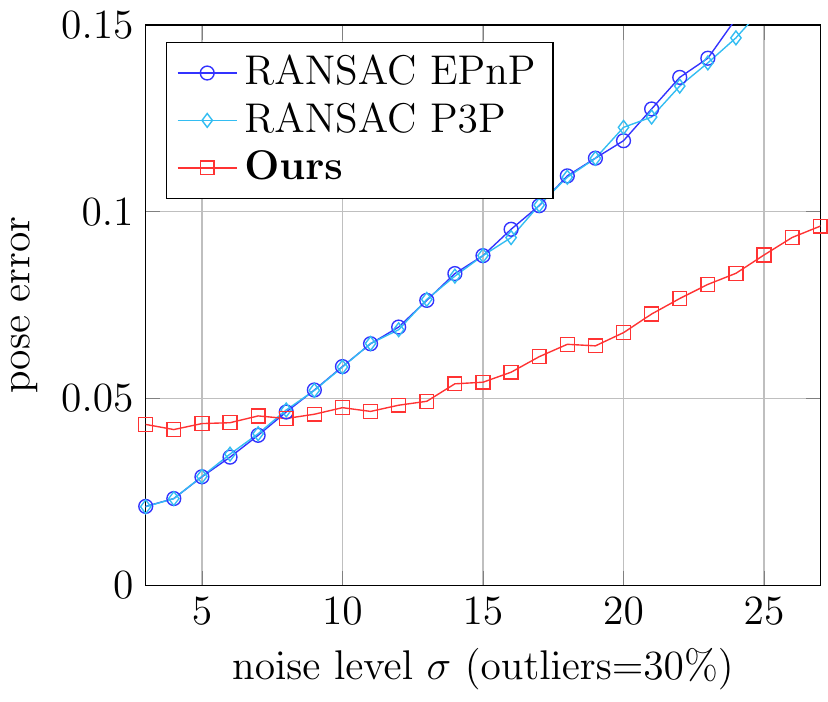}
    % \fbox{\rule{0pt}{1in} \rule{0.25\linewidth}{0pt}}
    \end{center}
    \vspace{-6mm}
    \caption{
        {\bf Comparison with RANSAC PnP.} We compare our network with two classical RANSAC-based PnP methods, EPnP~\cite{Lepetit09} and P3P~\cite{Gao03}. The two RANSAC-based methods have very similar performance. More importantly, our method is much more accurate and robust when the noise increases. The pose error is reported as the ratio of the 3D space reconstruction error to the diameter of the target object.
    }
    \label{fig:ours_vs_ransac_on_synthetic}
\end{figure}

{\bf Comparing with RANSAC PnP.} Combining a PnP algorithm with RANSAC is the most widespread approach to handling noisy correspondences~\cite{Rad17,Tekin18a,Hu19a,Zakharov19a}.  Fig.~\ref{fig:ours_vs_ransac_on_synthetic} shows that RANSAC-based EPnP~\cite{Lepetit09} and RANSAC-based P3P~\cite{Gao03} yield similar performance. While they are more accurate than our learning-based method when there is very little noise, our method quickly becomes much more accurate when the noise level increases.

\begin{figure}[t]
    \begin{center}
    \includegraphics[width=0.49\linewidth]{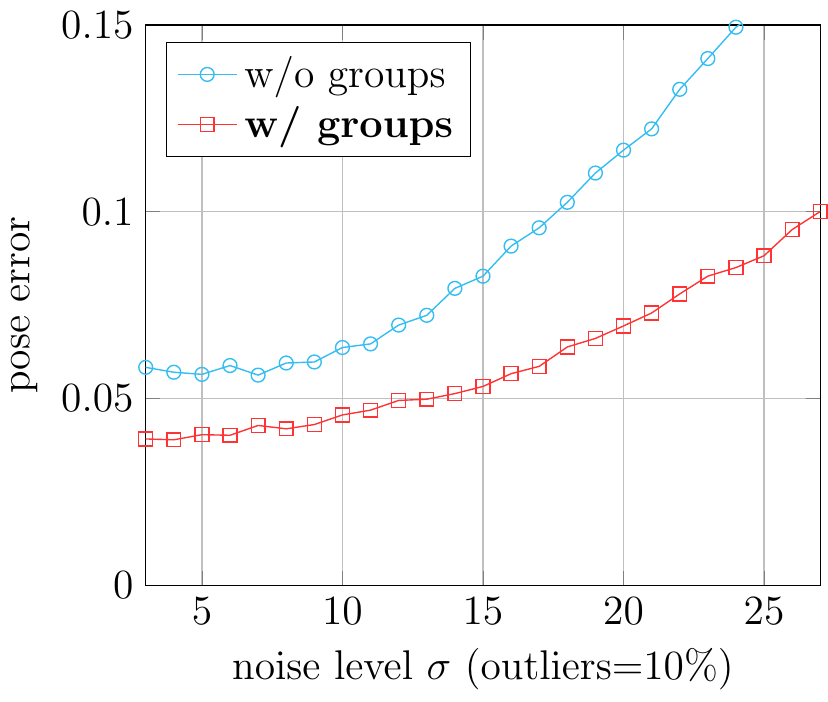}
    \includegraphics[width=0.49\linewidth]{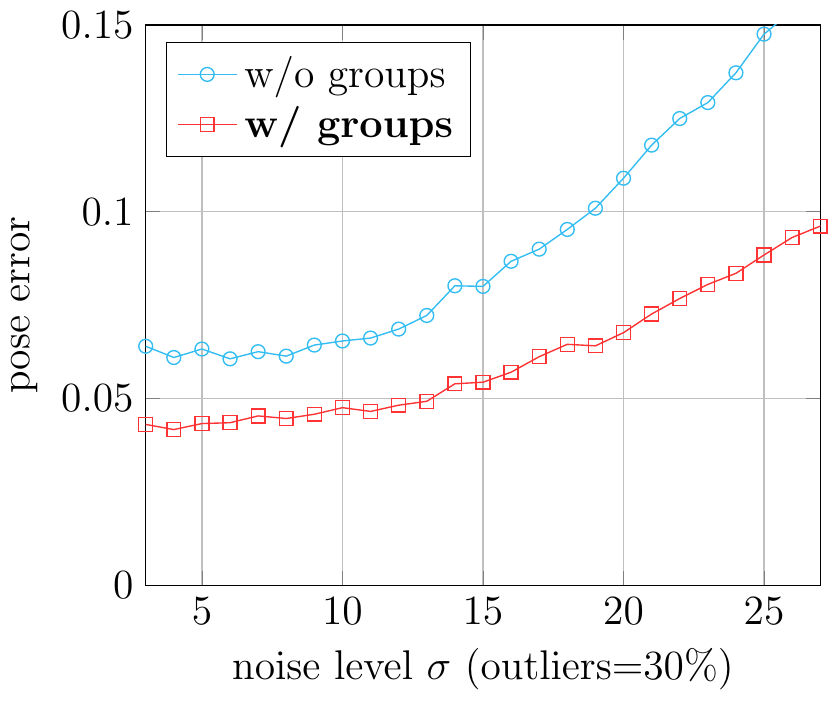}
    % \fbox{\rule{0pt}{1in} \rule{0.25\linewidth}{0pt}}
    \end{center}
    \vspace{-6mm}
    \caption{{\bf Importance of correspondence clustering.} We compare our network with one having a single max-pooling operation, thus not accounting for the order of the clusters. Ignoring this property clearly degrades the performance.
    }
    \label{fig:eval_wo_grouped}
\end{figure}

{\bf Importance of correspondence clustering.}
To showcase the importance to structure our network in the way we did, we implemented a simplified version that uses a single max-pooling operation  to achieve permutation invariance for all correspondences, without accounting for the order of the clusters that matches that of the keypoints. To make this work, we had to incorporate explicitly the 3D keypoint coordinates associated to each correspondence as input to the network. As shown in Fig.~\ref{fig:eval_wo_grouped}, not modeling the fixed order of the keypoints yields a significant decreases in accuracy.

\begin{figure}[t]
    \begin{center}
    \includegraphics[width=0.49\linewidth]{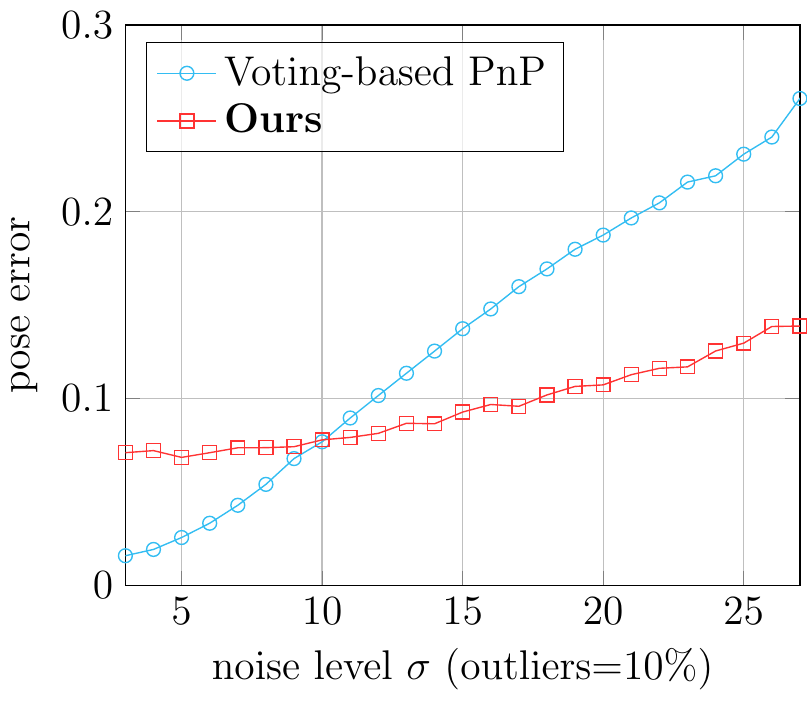}
    \includegraphics[width=0.49\linewidth]{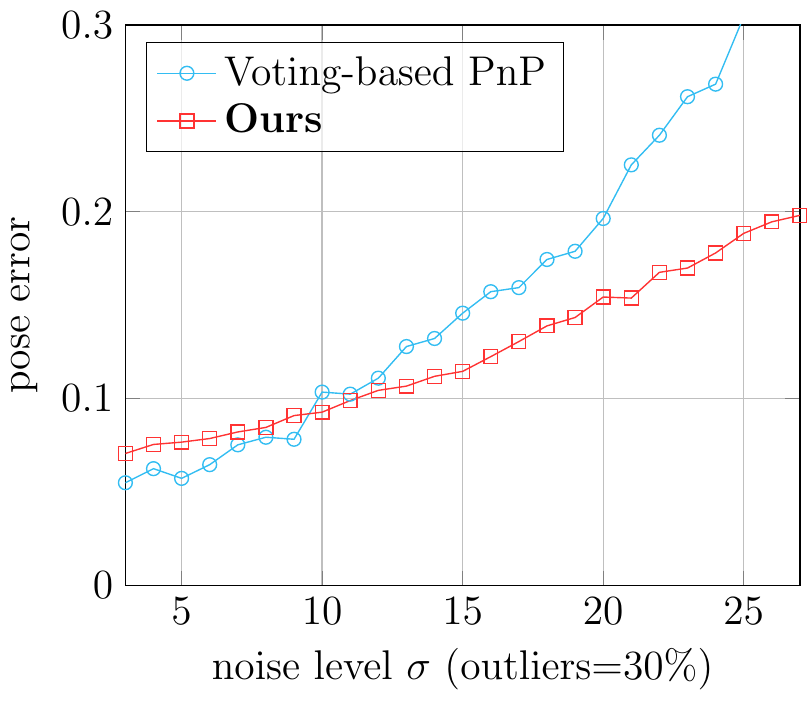}
    % \fbox{\rule{0pt}{1in} \rule{0.25\linewidth}{0pt}}
    \end{center}
    \vspace{-6mm}
    \caption{{\bf Comparison with PVNet's voting-based PnP~\cite{Peng19a}.} When using 3D point to 2D vector correspondences, we compare our network with the voting-based PnP used by PVNet. Our method is much more robust to noise than voting-based PnP.
    }
    \label{fig:ours_vs_pvnet_on_synthetic}
\end{figure}

{\bf Comparing with PVNet's voting-based PnP.} In the above experiments, the 2D correspondences were expressed in terms of 2D locations of image points. Since one of the best current techniques~\cite{Peng19a} uses directions instead and infers poses from those using a voting-based PnP scheme, we feed the same 3D point to 2D vector correspondences to our own network. In this setting, as shown in Fig.~\ref{fig:ours_vs_pvnet_on_synthetic}, the pose is more sensitive to the correspondence noise. However, as in the previous case, while voting-based PnP yields more accurate results when there is little noise, our method is much more robust and accurate when the noise level increases.

% !TEX root = ../top.tex
% !TEX spellcheck = en-US

% !TEX root = ../top.tex
% !TEX spellcheck = en-US

\subsection{Real Data}

We evaluate our method on real data from two challenging datasets, Occluded-LINEMOD~\cite{Krull15} and YCB-Video~\cite{Xiang18b}.

{\bf Occluded-LINEMOD} consists of 8 objects and is a subset of the older LINEMOD dataset~\cite{Hinterstoisser12b}. Unlike LINEMOD in which only one object per image is annotated, Occluded-LINEMOD features multiple annotated objects. This makes it more meaningful for evaluating methods that perform both instance detection and pose estimation. In addition to the cluttered backgrounds, textureless objects, and changing lighting conditions of LINEMOD, Occluded-LINEMOD also has severe occlusions between multiple object instances. As there are only 1214 testing images and no explicit training data in Occluded-LINEMOD, we train our network based on the LINEMOD training data.

{\bf YCB-Video} is more recent and even more challenging.  It features 21 objects taken from the YCB dataset~\cite{Calli15,Calli17} and comprises about 130K real images from 92 video sequences. It offers all the challenges of Occluded-LINEMOD plus more diverse object sizes, including several tiny textures-less objects.

% !TEX root = ../top.tex
% !TEX spellcheck = en-US

% !TEX root = ../top.tex
% !TEX spellcheck = en-US

\begin{table}
	\centering
	\rowcolors{3}{gray!10}{white}
	\begin{tabular}{lcc!{\vrule width 1pt}cc}
	\toprule
    & \cite{Hu19a} & \cite{Hu19a} + {\bf Ours}  & \cite{Peng19a} & \cite{Peng19a} + {\bf Ours}          \\
	\midrule                                  
	Ape			&12.1	& {\bf 14.8} &15.8	& {\bf 19.2}  \\
	Can			&39.9	& {\bf 45.5} &63.3	& {\bf 65.1}  \\
	Cat 		&8.2	& {\bf 12.1} &16.7	& {\bf 18.9}  \\
	Driller		&45.2	& {\bf 54.6} &65.7	& {\bf 69.0}  \\
	Duck		&17.2	& {\bf 18.3} &25.2	& {\bf 25.3}  \\
	Eggbox$^*$	&22.1	& {\bf 30.2} &50.2	& {\bf 52.0}  \\
	Glue$^*$	&35.8	& {\bf 45.8} &49.6	& {\bf 51.4}  \\
	Holepun.	&36.0	& {\bf 37.4} &39.7	& {\bf 45.6}  \\
	\midrule                                  
	Average		&27.0	&	{\bf 32.3} &40.8	& {\bf 43.3}  \\
	\bottomrule
	\end{tabular}
	\vspace{-2mm}
	\caption{{\bf Evaluation with different correspondence-extraction networks on Occluded-LINEMOD.} We evaluate two state-of-the-art correspondence-extraction networks: SegDriven~\cite{Hu19a} and PVNet~\cite{Peng19a}, by replacing their original RANSAC-based post processing with our small network. Our method consistently outperforms the original versions in both cases. Here, we report the ADD-0.1d.
    }
    \label{tab:ablation_on_linemod}
\end{table}

% !TEX root = ../top.tex
% !TEX spellcheck = en-US

\begin{table*}
	\centering
	\rowcolors{3}{white}{gray!10}
	\begin{tabular}{lcccc!{\vrule width 1pt}cccc}
	\toprule
	& \multicolumn{4}{c}{ADD-0.1d}	&	\multicolumn{4}{c}{REP-5px} \\
    &	PoseCNN & SegDriven & PVNet & {\bf Ours} &	PoseCNN & SegDriven & PVNet & {\bf Ours} \\
	\midrule
	Ape			&9.6	&12.1	&15.8	& {\bf 19.2} &34.6	&59.1	&69.1		& {\bf  70.3}\\
	Can			&45.2	&39.9	&63.3	& {\bf 65.1} &15.1	&59.8	&{\bf 86.1}	& 85.2\\
	Cat 		&0.9	&8.2	&16.7	& {\bf 18.9} &10.4	&46.9	&65.1		& {\bf  67.2}\\
	Driller		&41.4	&45.2	&65.7	& {\bf 69.0} &7.4	&59.0	&{\bf 73.1}	& 71.8\\
	Duck		&19.6	&17.2	&25.2	& {\bf 25.3} &31.8	&42.6	&61.4		& {\bf  63.6}\\
	Eggbox$^*$	&22.0	&22.1	&50.2	& {\bf 52.0} &1.9	&11.9	&8.4		& {\bf  12.7}\\
	Glue$^*$	&38.5	&35.8	&49.6	& {\bf 51.4} &13.8	&16.5	&55.4		& {\bf  56.5}\\
	Holepun.	&22.1	&36.0	&39.7	& {\bf 45.6} &23.1	&63.6	&69.8		& {\bf  71.0}\\
	\midrule                                         
	Average		&24.9	&27.0	&40.8	& {\bf 43.3} &17.2	&44.9	&61.1	& {\bf 62.3}\\
	\bottomrule
	\end{tabular}
	\vspace{-2mm}
	\caption{{\bf Comparison with the state of the art on Occluded-LINEMOD.} We compare our results with those of PoseCNN~\cite{Xiang18b}, SegDriven~\cite{Hu19a}, and PVNet~\cite{Peng19a} in terms of both ADD-0.1d and REP-5px. Our method outperforms the state of the art, especially in ADD-0.1d.
    }
    \label{tab:comparion_on_linemod}
\end{table*}

% !TEX root = ../top.tex
% !TEX spellcheck = en-US

\begin{figure*}[t]
    \begin{center}
    \includegraphics[width=0.24\linewidth, clip, trim=80 100 150 50]{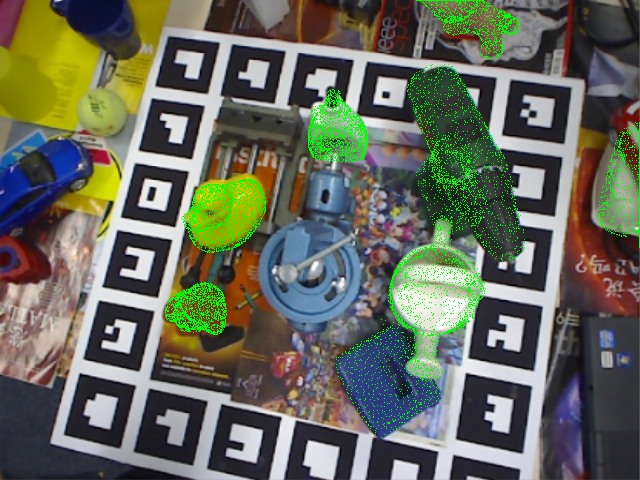} 
    \includegraphics[width=0.24\linewidth, clip, trim=80 100 150 50]{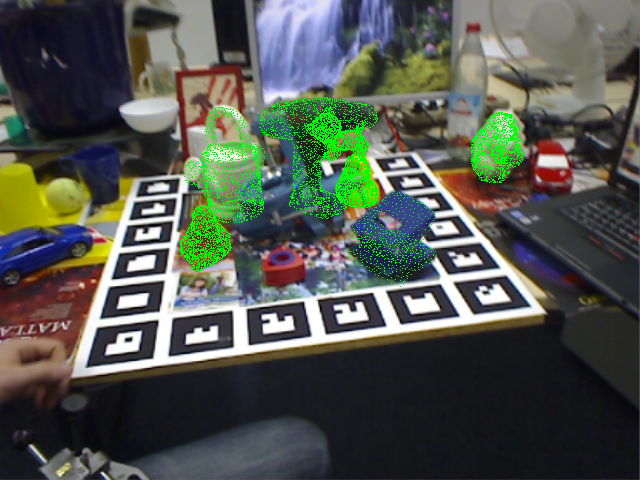} 
    \includegraphics[width=0.24\linewidth, clip, trim=50 150 180 0]{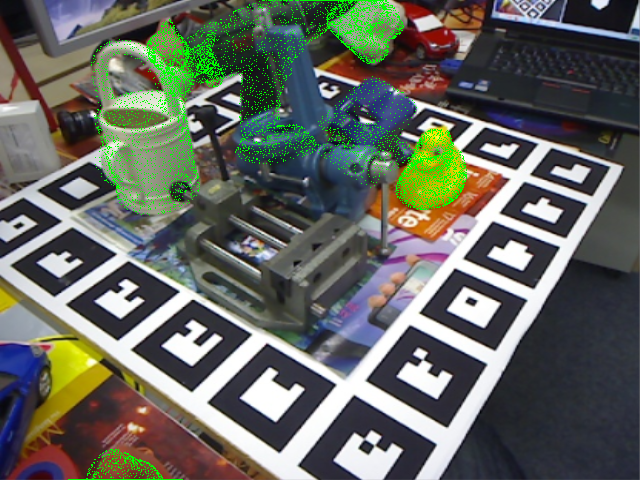} 
    \includegraphics[width=0.24\linewidth, clip, trim=230 150 0 0]{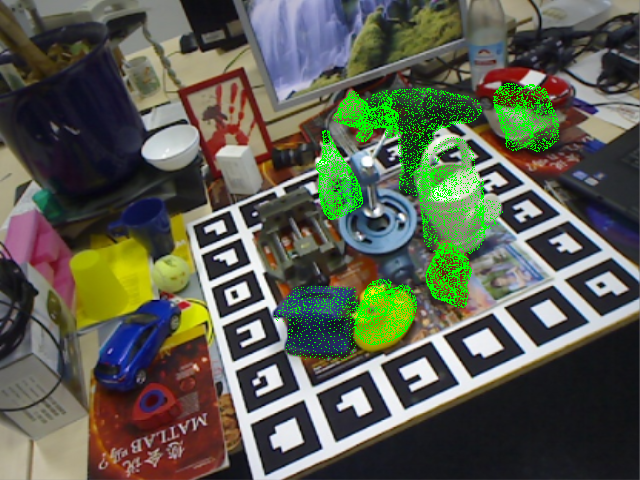} \\
    \vspace{0.5mm}
    \includegraphics[width=0.24\linewidth, clip, trim=80 50 150 100]{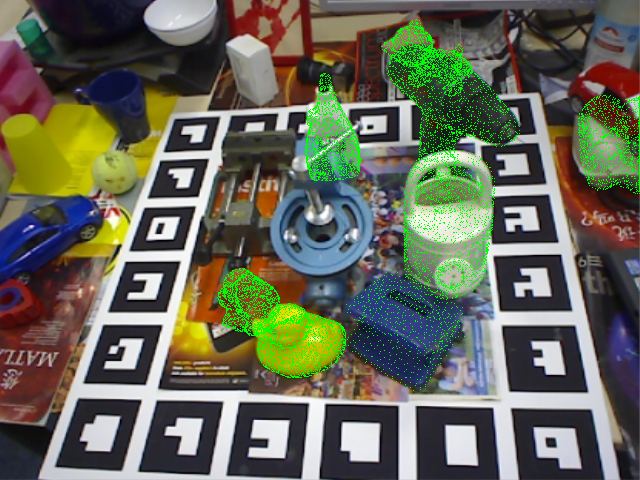} 
    \includegraphics[width=0.24\linewidth, clip, trim=150 150 80 0]{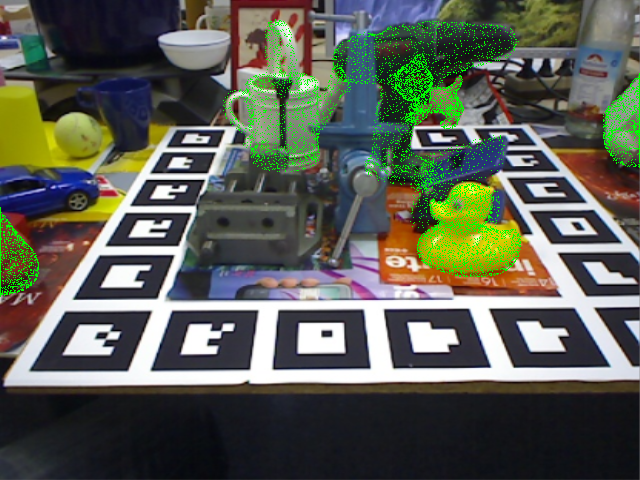} 
    \includegraphics[width=0.24\linewidth, clip, trim=180 100 50 50]{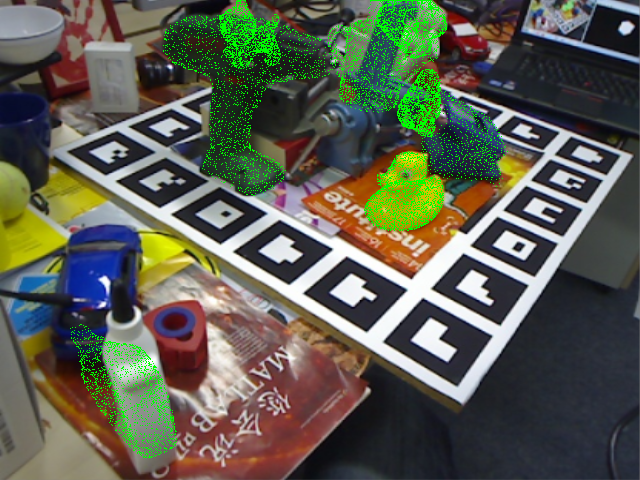} 
    \includegraphics[width=0.24\linewidth, clip, trim=0 0 230 150]{./fig/visual_demo/7.png} \\
    \end{center}
    \vspace{-6mm}
    \caption{{\bf Qualitative results on Occluded-LINEMOD}. Our method yields accurate results even in the presence of large occlusions, as shown in the first three columns. The last column shows two failure cases, where the target egg box is occluded too much and the target glue exhibits subtle symmetry ambiguities, making it not easy for the correspondence-extraction network~\cite{Peng19a} to establish stable correspondences. Here, the pose is visualized as the reprojection of the 3D mesh for each object.
    }
    \label{fig:visual_demo}
\end{figure*}

% !TEX root = ../top.tex
% !TEX spellcheck = en-US

\begin{table}
	\centering
	\scalebox{0.9}{
	% \rowcolors{2}{white}{gray!10}
	\begin{tabular}{lC{8em}cC{3em}c}
	\toprule
    &	correspondence \newline extraction & fusion & total time & FPS \\
	\midrule
	% Speed (FPS) & 4 & 22 & 25 & {\bf 43}\\
	PoseCNN 	& -		&	-	&	$>$250	& $<$4\\
	SegDriven 	& 30	& 20	& 50 & 20\\
	PVNet 		&  {\bf 14}	& 26  & 40 & 25\\
	{\bf Ours} 	& {\bf 14} & {\bf 8}& {\bf 22} & {\bf 45}\\
	\bottomrule
	\end{tabular}
	}
	\vspace{-2mm}
	\caption{{\bf Comparing speed.} We compare the running times (in milliseconds) of PoseCNN~\cite{Xiang18b}, SegDriven~\cite{Hu19a}, PVNet~\cite{Peng19a} and our method on a modern GPU (GTX1080 Ti). Except for PoseCNN, these methods first extract correspondences and then fuse them. With the same correspondence-extraction backbone as in PVNet, our method runs about 2 times faster, thanks to our network that prevents the need for RANSAC-based fusion.
    }
    \label{tab:speed_on_linemod}
\end{table}

% !TEX root = ../top.tex
% !TEX spellcheck = en-US

\begin{table}
	\centering
	% \rowcolors{2}{white}{gray!10}
	\begin{tabular}{lcccc!{\vrule width 1pt}cccc}
	\toprule
	& ADD-0.1d	& REP-5px \\
	\midrule 
	PoseCNN & 21.3 & 3.7 \\
	SegDriven & 39.0 & 30.8 \\
	PVNet & - & 47.4 \\
	{\bf Ours} & {\bf 53.9} & {\bf 48.7}\\
	\bottomrule
	\end{tabular}
	\vspace{-2mm}
	\caption{{\bf Comparison with the state of the art on YCB-Video.} We compare our results with those of PoseCNN~\cite{Xiang18b}, SegDriven~\cite{Hu19a}, and PVNet~\cite{Peng19a} in terms of ADD-0.1d and REP-5px. We denote by ``-'' the result missing from the original PVNet paper.}
    \label{tab:comparion_on_ycb}
\end{table}

\parag{Data preparation.}
For Occluded-LINEMOD, as in~\cite{Tekin18a,Hu19a,Peng19a}, we first use the Cut-and-Paste synthetic technique~\cite{Dwibedi17a} to generate 20K images from LINEMOD data and random background data~\cite{Xiao10b}, with 4 to 10 different instances for each image. Then, we generate 10K rendering images for each object type from the textured 3D mesh, as in~\cite{Peng19a}. The pose range during the rendering procedure is  the same as in LINEMOD except for one thing: To handle pose ambiguities when encountering symmetry objects~\cite{Manhardt19}, we restrict the pose range to a subrange according to the symmetry type of the object during training to avoid confusing the network~\cite{Rad17}. In the end, our training data consists of 20K synthetic images with multiple instances and 10K rendered images with only one instance for each object, a total of ($20+10\times 8$)K images.

For YCB-Video, we follow a similar procedure. We render 10K images for each of the 21 objects using the 3D mesh models that are provided and according to the pose statistic of the dataset. However, we do not use the Cut-and-Paste technique to generate images with multiple instances because in the original YCB-Video images are already annotated with multiple objects and we use that directly.

\parag{Training Procedure.}
For both datasets, we use an input image resolution of 640 $\times$ 480 for both training and testing, as in~\cite{Peng19a}. We use Adam to optimize with the initial learning rate set to 1e-3 and divided by 10 after processing 50\%, 75\%, and 90\% of the total number of data samples. We set the batch size to be 8 and rely on the usual data augmentation techniques, that is as random luminance, Gaussian noise, translation, scaling, and also occlusions~\cite{Zhong17}. We train the network on 5M training samples through online data augmentation.

\parag{Metrics.}
We quantify the pose error in both 3D and 2D as in~\cite{Xiang18b,Hu19a}. In 3D, it use the average distance between the 3D model points transformed using the predicted pose and those obtained with the ground-truth one, and we refer to it as ADD~\cite{Xiang18b}. In 2D, we use
the usual 2D reprojection error of the 3D model points, and we refer it as REP~\cite{Hu19a}. We measure the pose accuracy in terms the percentage of recovered poses that are correct. In the tables below, we report ADD-0.1d and REP-5px, for which the predicted pose are considered to be correct if ADD is smaller than 10\% of the model diameter and REP is below 5 pixel, respectively. For each metric, we use the symmetric version for symmetric objects, which we denote by a $^*$ superscript.

\subsubsection{Occluded-LINEMOD Results}
\label{sec:eval_occlinemod}

As discussed before, to demonstrate that our method is generic, we test it in conjunction with  two correspondence-extraction networks SegDriven~\cite{Hu19a} and PVNet~\cite{Peng19a}. Table~\ref{tab:ablation_on_linemod} shows that, by replacing the original RANSAC-based post processing by our network to turn the approach into a single-stage one we improve performance in both cases. 

In Table~\ref{tab:comparion_on_linemod}, we shown that our single-stage network outperform the state-of-the-art methods, PoseCNN~\cite{Xiang18b}, SegDriven~\cite{Hu19a} and PVNet~\cite{Peng19a}. Fig.~\ref{fig:visual_demo} provides qualitative results.  In Table~\ref{tab:speed_on_linemod}, we report runtimes for an input image containing about 4-5 objects. Our method is also faster than the others
%, mainly 
because it does away for the iterative RANSAC procedure. 
% \MS{I removed the "mainly", because it is truly the reason (the backbone is the same).}

\subsubsection{YCB-Video Results}
\label{sec:eval_ycb}

Table~\ref{tab:comparion_on_ycb} summarizes the results comparing against PoseCNN~\cite{Xiang18b}, SegDriven~\cite{Hu19a}, and PVNet~\cite{Peng19a}. It shows that our method consistently also outperforms the others on this dataset. Furthermore, note that it runs nearly 10 times faster than PoseCNN and also nearly 2 times faster than SegDriven and PVNet. 

\subsection{Limitations}
\label{sec:discuss}

While our method is accurate and fast when used in conjunction with state-of-the-art correspondence-extraction networks~\cite{Hu19a,Peng19a}, the network that estimates the poses from the correspondences is still not as accurate as traditional geometry-based PnP algorithms when very precise correspondences can be obtained by other means, as shown in Fig.~\ref{fig:ours_vs_ransac_on_synthetic}. Furthermore, it does not address the generic PnP problem because we only trained it for fixed sets of 3D coordinates. Addressing this will be the focus of our future work.

% !TEX root = ../top.tex
% !TEX spellcheck = en-US

\section{Conclusion}
\label{sec:conclusion}

We have introduced a single-stage approach approach to 6D detection and pose estimation. Its key ingredient is a small network that takes candidate 3D-to-2D correspondences and returns a 6D pose. When combined with state-of-the-art approaches to establish the correspondences, it boosts performance by allowing end-to-end training and eliminating the cumbersome RANSAC style procedure that they normally require. 

Future work will focus on making the pose estimation network more accurate and more generic so that it can be used in a broader context. 
 
\subsubsection*{Acknowledgments}
This work was supported in part by the Swiss Innovation Agency (Innosuisse). We would like to thank Vincent Lepetit, Kwang Moo Yi and Eduard Trulls for helpful discussions.

{\small
\bibliographystyle{ieee_fullname}
\bibliography{string,graphics,vision,learning}
}

\end{document}